\documentclass{article}
\usepackage{spconf,amsmath,graphicx}
\usepackage{multirow}
\usepackage{booktabs}
\usepackage{amssymb}

\title{CluCDD: Contrastive Dialogue Disentanglement via Clustering}
\name{Jingsheng Gao, Zeyu Li, Suncheng Xiang, Ting Liu, Yuzhuo Fu} 

\address{School of Electronic Information and Electrical Engineering, Shanghai Jiao Tong University, China}
\begin{document}
%
\maketitle
\begin{abstract}
A huge number of multi-participant dialogues happen online every day, which leads to difficulty in understanding the nature of dialogue dynamics for both humans and machines. Dialogue disentanglement aims at separating an entangled dialogue into detached sessions, thus increasing the readability of long disordered dialogue. Previous studies mainly focus on message-pair classification and clustering in two-step methods, which cannot guarantee the whole clustering performance in a dialogue. To address this challenge, we propose a simple yet effective model named \textbf{CluCDD}, which aggregates utterances by contrastive learning. More specifically, our model pulls utterances in the same session together and pushes away utterances in different ones. Then a clustering method is adopted to generate predicted clustering labels. Comprehensive experiments conducted on the Movie Dialogue dataset and IRC dataset demonstrate that our model achieves a new state-of-the-art result\footnote{ Code is available at  https://github.com/gaojingsheng/CluCDD}.
\end{abstract}
\section{Introduction}
\label{sec:intro}
\begin{figure}[!t]
	\centering
	\includegraphics[width=0.45\textwidth]{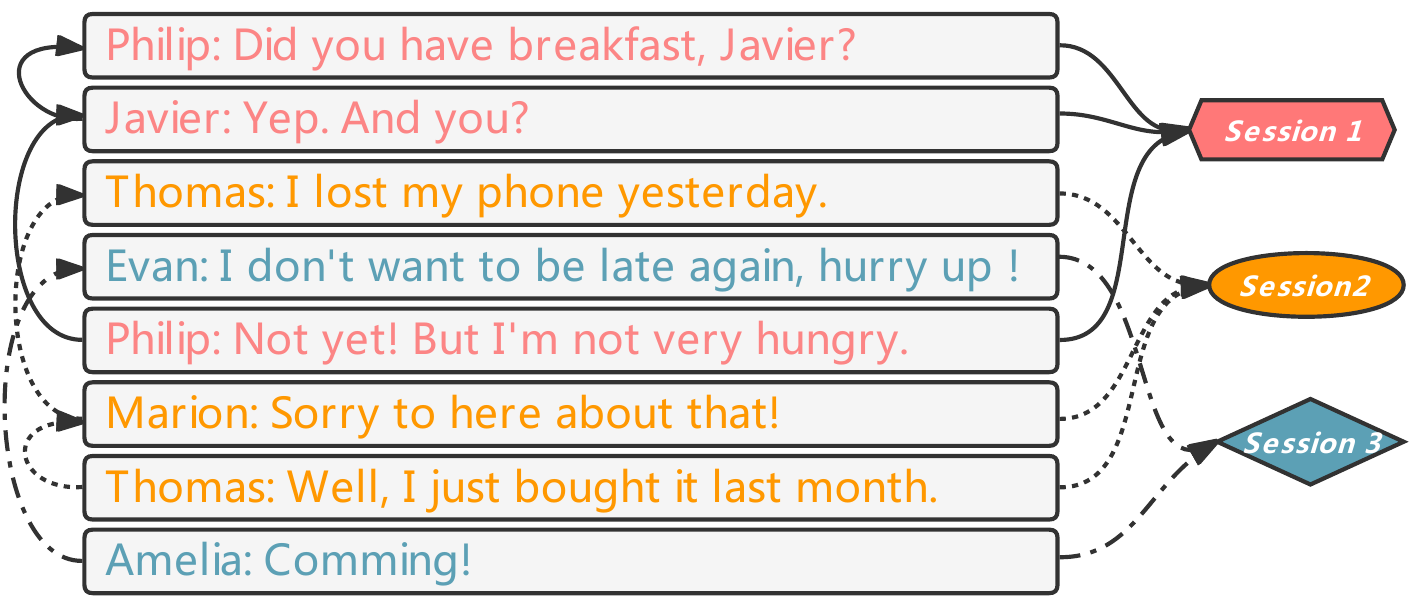}
	\caption{An example of dialogue disentanglement.}
	\label{fig:sessions}
\end{figure}

With the rapid development of the Internet and chatting apps, online group chatting has increased popularity, which also generates many multi-party dialogues~\cite{lowe-etal-2015-ubuntu}. In many cases, multi-party dialogues include many users, and every user's messages are entangled with each other, making it difficult for a new user to grasp the previous topics quickly. As shown in Fig \ref{fig:sessions}, several sessions compose a larger dialogue randomly. This kind of entanglement will bring difficulty for new users to find a specific topic. Automatic dialogue disentanglement will segment entangled utterances into different sessions and help users and machine find a specific session quickly.

Owing to the automatical topics separating, dialogue disentanglement is proved to be valuable in handling the corresponding downstream tasks~\cite{dad.2017.102,jia-etal-2020-multi,mpcsuvey,ouchi2016addressee}. Given the considerable variability in the content of each dialogue, a conventional classification method is not readily applicable to the task of disentangling dialogues. Existing methods can be roughly divided into two categories: two-step and end-to-end. Two-step methods~\cite{mehri-carenini-2017-chat,CISIR} get the reply relationships among message pairs firstly, then apply a clustering method to establish different sessions based on the message pairs relationships. However, these two-step methods are susceptible to noisy utterance pair relations, resulting in poor final clustering results. End-to-end methods~\cite{ijcai2020-535, liu2021unsupervised} are proposed to fill the gap between two steps and usually perform better than two-step methods, where the dialogue and session representations can be directly used to predict the clustering results. Liu et al~\cite{ijcai2020-535} proposed an end-to-end transition-based model in a supervised way, and their E2E model classifies an utterance into an existing or a new session. Besides, Liu et al~\cite{liu2021unsupervised} proposed an unsupervised co-training method based on the pseudo data generated by speaker labels. However, previous end-to-end methods have not considered aggregating utterances in one dialogue directly, and they mainly focus on classifying the relations between utterances and sessions. \\
Recently, Pre-trained Language Models (PrLMs) have considerably improved downstream natural language process tasks by providing effective backbones. Based on the pre-trained BERT~\cite{devlin-etal-2019-bert}, we construct an end-to-end framework: \textbf{C}ontrastive \textbf{D}ialogue \textbf{D}isentanglement via \textbf{Clu}stering (\textbf{CluCDD}). Our approach focuses on distinguishing utterances in different sessions for entangled dialogues by contrastive learning. We first retrieve the utterances representations in each dialogue through pre-trained BERT. Since utterances in dialogues are temporally coherent with preceding utterances, we capture the sequential information by a sequential feature fusion (SFF) encoder. Meanwhile, the number of clusters is required in some clustering methods, so we put forward a cluster head to predict the session number for clustering. Finally, we forge our session outcomes through a clustering process.
\begin{figure*}[!t]
	\centering
	\includegraphics[width=0.85\textwidth]{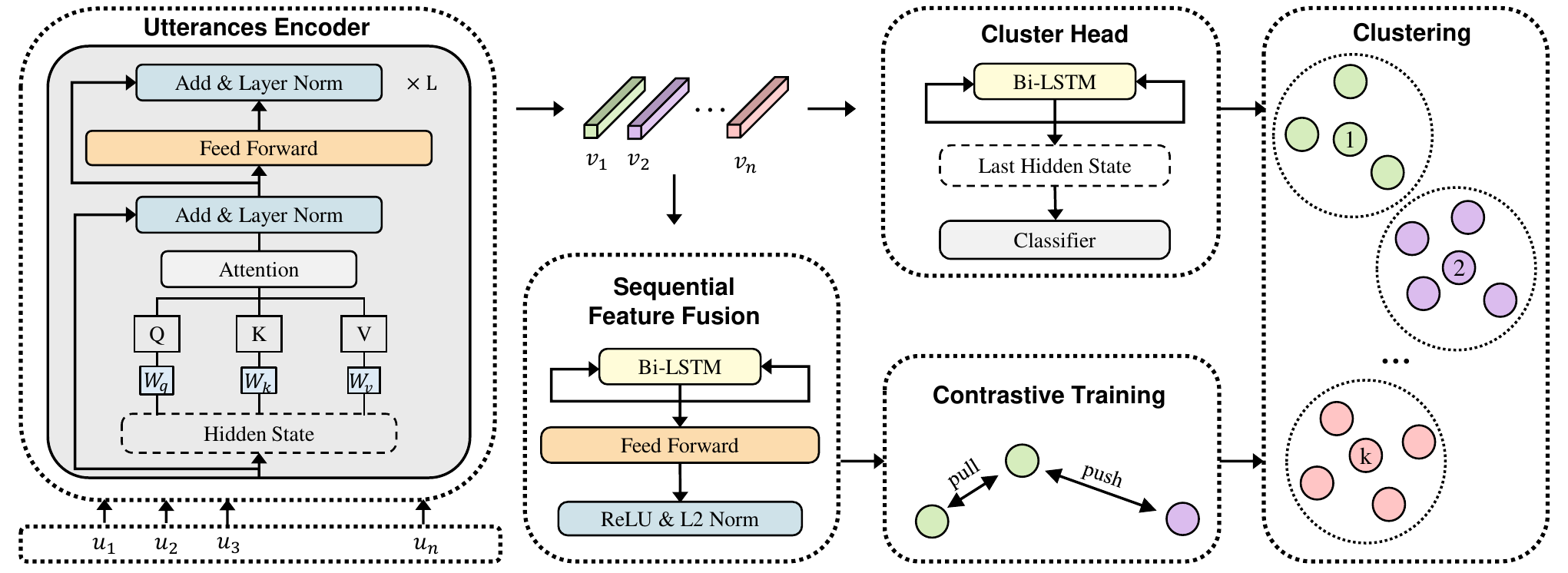}
	\caption{The architecture of our CluCDD. (a) We use BERT to encode the utterances in one dialogue, all utterances share the same parameters. (b) The utterances representations are fed into the SFF module retrieve the sequential features. (c) We adopt a Cluster Head to predict session number k. (d) We generate predicted labels by a clustering method for utterances. }
	\label{fig:model}
\end{figure*}
We conduct experiments on the Movie Dialogue dataset~\cite{ijcai2020-535} and Ubuntu IRC dataset~\cite{kummerfeld-etal-2019-large}, and results show that our CluCDD outperforms the existing methods. Our contributions are summarized as follows:
\begin{itemize}
	\item We propose CluCDD, an effective framework for dialogue disentanglement. The model captures the utterances and sequential representations in dialogues.
	\item The contrastive training paradigm is employed to amend feature space, and a cluster head predicts the session number to enhance the final clustering result.
	\item Extensive experiments demonstrate that our CluCDD is suitable for solving dialogue disentanglement and establishes the state-of-the-art on two datasets. 
\end{itemize}

\section{Proposed Methodology}
\label{sec:Methodology}

\subsection{Problem Definition}

Dialogue disentanglement is a clustering task. Given a dialogue $D={u_1, u_2, \cdots, u_n}$, where $u_i$ represents the $i$-th utterance in $D$. An utterance in a dialogue includes a session number $l_i$, and different sessions commonly mean different topics or reply-to relations. In this task, we aim to separate the dialogue $D$ into different sessions $l_1, l_2, \cdots, l_k$, where session $l_i$ contains $m_i$ utterances and $\sum_{i=1}^{k} m_i = n$. 

\subsection{Model Architecture}

\textbf{Utterance Encoder}. BERT~\cite{devlin-etal-2019-bert} yields strong performances across many downstream natural language processing tasks. In CluCDD, we employ pre-trained BERT as our utterance encoder. BERT is consisted of twelve($L$) blocks, where each block contains two types of sub-layers: multi-head self-attention and a fully connected feed-forward network. We assume that $\boldsymbol{u}_i$ represent a certain utterance in dialogue D, the input of $\boldsymbol{u}_i$ is processed into $[\operatorname{CLS}, \operatorname{word_1}, \operatorname{word_2}, ... \operatorname{word_m}]$, and $\operatorname{CLS}$ is the embedding for pre-trained classified task: 
\begin{equation}
	\boldsymbol{u}_{i,1}, \boldsymbol{u}_{i,2}, \cdots,\boldsymbol{u}_{i,m+1} = \operatorname{BERT} (\boldsymbol{u}_i)
\end{equation}
where $\mathbf{u}_{i,k}$ represent $k$-th output embedding derived from $u_i$. $\forall j=1,2,..,m+1$,  $\boldsymbol{u}_{i,j} \in \mathbb{R}^{1 \times d}$, where $m$ is the words number of the utterance and $d$ is the output dimension of BERT. Following previous work~\cite{zhang2021discovering}, we utilize the mean-pooling to get the average utterance embedding from BERT. \\
\noindent \textbf{Sequential Feature Fusion (SFF)}. Utterance's chronological order is significant to its semantic meaning in a dialogue. Hence, the historical utterances and the content of the dialogue are momentous to the meaning of a certain utterance. Thus we apply a sequential feature fusion module to capture the temporal context information. 

A full-connected (FC) layer is added after the output of BERT, then the $i$-th utterance representation $\boldsymbol{v}_i$ is:
\begin{equation}
	\boldsymbol{v}_i = \boldsymbol{W}(f(\boldsymbol{u}_{i,1}, \boldsymbol{u}_{i,2}, \cdots,\boldsymbol{u}_{i,m+1}))+\boldsymbol{b}
	\label{funw}
\end{equation}
where $f$ represents the mean-pooling layer, $\boldsymbol{W}$ and $\boldsymbol{b}$ are parameters of the FC layer, and $\boldsymbol{W} \in \mathbb{R}^{d \times d}$,  $\boldsymbol{b}\in \mathbb{R}^{1 \times d}$. Then we put a Bi-LSTM~\cite{LSTM} layer for compensating contextual clues within the whole dialogue.

The input of the Bi-LSTM is all $n$ utterances representations ${\boldsymbol{v}_1, \boldsymbol{v}_2, \cdots, \boldsymbol{v}_n}$ in one dialogue. Then all utterances representations after Bi-LSTM can be defined as:
\begin{equation}
	\boldsymbol{h}_1, \boldsymbol{h}_2, \cdots, \boldsymbol{h}_n  = \operatorname{Bi-LSTM} (\boldsymbol{v}_1, \boldsymbol{v}_2, \cdots, \boldsymbol{v}_n)
\end{equation}

Given the previous utterances representations with sequential information, we would like to perform a regularization process to prevent overfitting and extreme cases. The other important sublayer is the fully connected feed-forward network which consists of a linear transformation with a ReLU activation function in between:
\begin{equation}
	\boldsymbol{r}_1, \boldsymbol{r}_2, \cdots, \boldsymbol{r}_n  = \Psi(\operatorname{ReLU}(\boldsymbol{W}(\boldsymbol{h}_1, \boldsymbol{h}_2, \cdots, \boldsymbol{h}_n)+\boldsymbol{b}))
\end{equation}
where $\Psi$ represents the L2 Normalization and $\boldsymbol{W} \in \mathbb{R}^{d \times d}$,  $\boldsymbol{b}\in \mathbb{R}^{1 \times d}$. \\
\noindent \textbf{Cluster Head}. Cluster number is a significant parameter in several clustering methods, e.g. K-means~\cite{kmeans}. To address this problem, we add an extra cluster head to predict the session number in each dialogue. Our cluster head comprises an LSTM layer and a linear layer, which share the same input as the SFF module. The training loss of our cluster head $\mathcal{L}_{H}$ is the cross-entropy loss, which can be formulated as:
\begin{equation}
	\label{eq:head_loss}
	\mathcal{L}_{H} = \mathbb{E}_{y \sim P}[-\log P(y=k)]
\end{equation}
where $k$ is the golden session number for each dialogue.
\subsection{Contrastive Learning}

To enforce maximizing the consistency between positive pairs compared with negative pairs, we adopt contrastive learning for our clustering methods. Consider ${\mathbf{r}_{1}, \mathbf{r}_{2}, \cdots, \mathbf{r}_{n}}$ are the output of SFF module. Let $\mathbf{r}_{i}, \mathbf{r}_{j}$ be a pair of input vectors, $y_{ij}$ be a binary label assigned to this pair. $y_{ij}=0$ if $\mathbf{r}_{i}$ {and} $\mathbf{r}_{j}$ are deemed similar, $y_{ij}=1$ if $\mathbf{r}_{i}$ and $\mathbf{r}_{j}$ are deemed dissimilar.

Following ~\cite{hadsell2006dimensionality}, we adopt a euclidean distance-based loss function for better performance. The distance function to be learned $D_{W}$ between $\mathbf{r}_i$ and $\mathbf{r}_j$ as the euclidean distance between the outputs of $G_{W}$. It can be formulated as:
\begin{equation}
	D_{W}^{ij}=D_{W}\left(\mathbf{r}_i, \mathbf{r}_j \right)=\left\|G_{W}\left(\mathbf{r}_i\right)-G_{W}\left(\mathbf{r}_j\right)\right\|_{2}
\end{equation}
Consequently, we can retrieve the contrastive loss function:
\begin{equation}
	\mathcal{L}_{C}=\sum_{i=1}^{n}\sum_{j=i+1}^{n} \mathcal{L}_c\left(\mathbf{r}_i, \mathbf{r}_j, y_{ij}\right)
\end{equation}
\begin{equation}
	\mathcal{L}_c\left(\mathbf{r}_i, \mathbf{r}_j, y_{ij}\right)=(1-y)\mathcal{L}_S(D_{W}^{ij})+y\mathcal{L}_D(D_{W}^{ij})
\end{equation}
$\mathcal{L}_{S}$ and $\mathcal{L}_{D}$ are designed for minimizing $\mathcal{L}$, where $\mathcal{L}_{S}$ for similar pairs while $\mathcal{L}_{D}$ for dissimilar pairs.

In our experiment, we set the exact loss function:
\begin{equation}
	\begin{aligned}
		&\mathcal{L}_c\left(r_i, r_j, y_{ij})\right)= \\
		&\quad(1-y_{ij}) \frac{1}{2}\left(D_{W}^{ij}\right)^{2}+(y_{ij}) \frac{1}{2}\left\{\max \left(0, m -D_{W}^{ij}\right)\right\}^{2}
	\end{aligned}
	\label{eq:marign}
\end{equation}
where margin $m>0$ is a hyper-parameter which can prevent loss from being less than zero.

Adding up the $\mathcal{L}_{H}$ in equation \ref{eq:head_loss}, we have the total training loss $\mathcal{L}$:
\begin{equation}
	\label{eq:loss}
	\mathcal{L} = \mathcal{L}_{C} + \gamma \mathcal{L}_{H}
\end{equation}

\section{Experiments}

\begin{table}
	\centering
	\footnotesize
	\setlength{\tabcolsep}{1.5mm}
	\caption{\label{Result} The results of dialogue disentanglement on the Movie Dialogue(MD) dataset and Ubuntu IRC dataset. SFF represents the Sequential Feature Fusion module. BL represents the Bi-LSTM layer in the SFF module.}
	\begin{tabular}{ccccccc}
		\toprule[1pt]
		Dataset & Method & NMI &  ARI & Lo$\textbf{c}_{3}$  & 1-1 & Shen-F \\
		\midrule
		\multirow{6}{*}{MD} & CISIR~\cite{CISIR} & 20.47 & 6.45 & - & - & 53.77 \\
		& BERT~\cite{devlin-etal-2019-bert} & 25.57 & 10.97 &- &- &  56.91 \\
		& E2E~\cite{ijcai2020-535} & 35.30 & 24.90 & - &- &  64.7 \\ \cmidrule{3-7}
		& CluCDD w/o SFF & 37.25 & 27.13 & 63.46 & 57.53 &  65.06 \\
		& CluCDD w/o BL   & 38.57 & 28.95   &  64.29 & 58.14 & 65.55 \\
		&CluCDD & \textbf{40.98} & \textbf{31.45} &  \textbf{67.98} & \textbf{61.75} &  \textbf{67.92} \\
		\midrule
		\multirow{6}{*}{IRC} & CISIR~\cite{CISIR} & 46.62 & 3.37 & - & - &  40.78\\
		& BERT~\cite{devlin-etal-2019-bert} & 54.61 & 8.15 & -&- &  43.87 \\	
		& E2E~\cite{ijcai2020-535} & 61.4 & 18.00 & - & - & 48.19 \\ \cmidrule{3-7}
		& CluCDD w/o SFF & 54.47 & 14.68 & 61.07 & 41.52 & 49.97  \\
		& CluCDD w/o BL   & 58.83   &  18.16  & 61.15 &  44.08 &  51.62 \\
		&CluCDD & \textbf{64.98} & \textbf{28.36} & \textbf{61.52} & \textbf{51.14}  & \textbf{58.42}  \\
		\bottomrule
	\end{tabular}
	
\end{table}

\subsection{Dataset and Training details}

We conduct experiments on two dialogue disentanglement datasets: Movie Dialogue dataset~\cite{ijcai2020-535} and IRC dataset~\cite{kummerfeld-etal-2019-large}. The larger Movie Dialogue dataset is collected from online movie scripts, which contains 29669/2036/2010 instances for train/dev/test. The origin label of an utterance in the Movie Dialogue dataset is session label, so it can be used for end-to-end dialogue disentanglement directly, and the session number of one dialogue is 2, 3 or 4, respectively.

The IRC dataset is annotated from online conversations, whose labels are the reply-to relations between utterances pairs. For the reason of lacking direct annotations of session labels for utterances in dialogues, the IRC dataset is used for two-step ways for dialogue disentanglement originally. Similar to~\cite{ijcai2020-535}, we process every continued 50 utterances into a dialogue. The minimum and maximum session numbers in our generated IRC dataset are 2 and 14.
We adopt the Adam optimizer with the initial learning rate of 5e-4. The hidden size of all layers is set to be 768. Meanwhile, as suggested in~\cite{zhang2021discovering}, we freeze all but the last transformer layer parameters of pre-trained BERT to speed up the training procedure and improve the training availability. The $ \gamma $ in equation \ref{eq:loss} is set to be 0.1.

\subsection{Comparison with the State-of-the-art}

Following previous work~\cite{ijcai2020-535,liu2021unsupervised,kummerfeld-etal-2019-large} on dialogue disentanglement, five evaluation metrics are employed in our experiments to evaluate the performance of different methods: Normalized Mutual Information (NMI), Adjusted Rand Index (ARI),  Lo$c_3$, One-to-one Overlap (1-1) and Shen F value (Shen-F). All of these measures with higher scores imply more accurate clustering results. 

The results in the Movie Dialogue dataset and the Ubuntu IRC dataset are presented in Table \ref{Result}, where the best results are highlighted in bold. CISIR is a two-step method proposed by \protect~\cite{CISIR}. BERT means retrieving the relationships between utterance pairs, then applying the clustering method by \protect~\cite{CISIR} to retrieve the results. E2E is an end-to-end method proposed by \protect~\cite{ijcai2020-535}, which can predict the session labels directly. CluCDD w/o SFF denotes that we fine-tune the pre-trained BERT in contrastive manner, then apply K-means and session number k retrieved by cluster head to generate the results. CluCDD w/o BL represents that we remove the Bi-LSTM layer in the SFF module. CluCDD is our proposed model, and the clustering method used in Table \ref{Result} is also K-means. 

From Table \ref{Result}, we can observe that our CluCDD outperforms all baselines on the Movie Dialogue dataset and IRC dataset. Besides, Bi-LSTM is significant in capturing the sequential information between utterances since each utterance concept can retain the hidden state of the whole dialogue.

We attribute these improvements on CluCDD to three main reasons: First, the direct clustering method is suitable for solving dialogue disentanglement, which is a clustering problem. Second, contrastive learning makes utterances in the same session become closer and utterances in different sessions become further, which is significant in separating an entangled dialogue into several sessions. Third, CluCDD can make full use of pre-trained knowledge and utilize the utterances representations and sequential feature extraordinarily.

\subsection{Ablation study}

\begin{figure}[ht]
	\centering
	\begin{minipage}[t]{0.52\linewidth}
		\centering
		\includegraphics[width=0.99\textwidth]{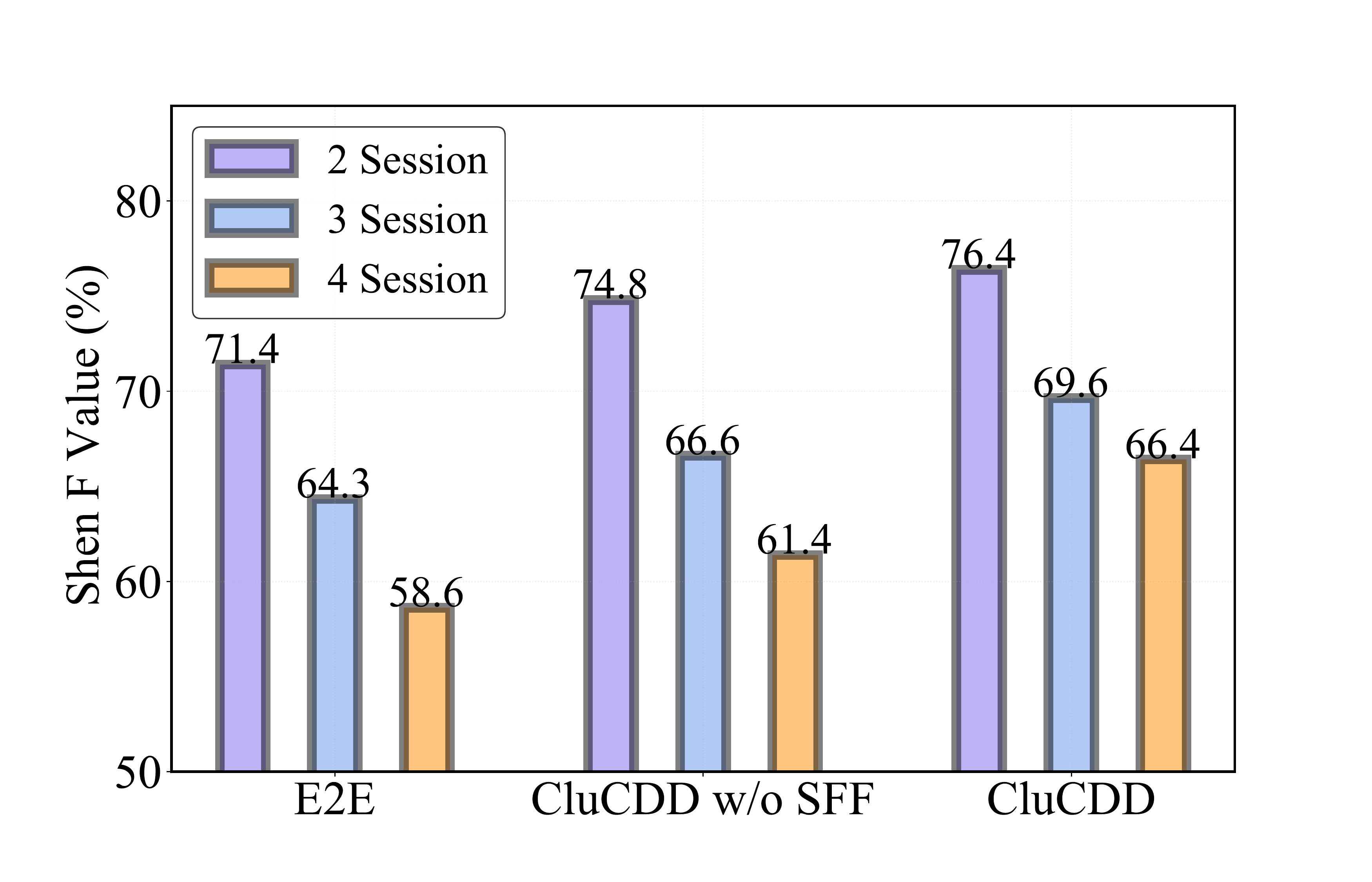}
		\label{fig52}
	\end{minipage}
	\begin{minipage}[t]{0.47\linewidth}
		\centering
		\includegraphics[width=0.98\textwidth]{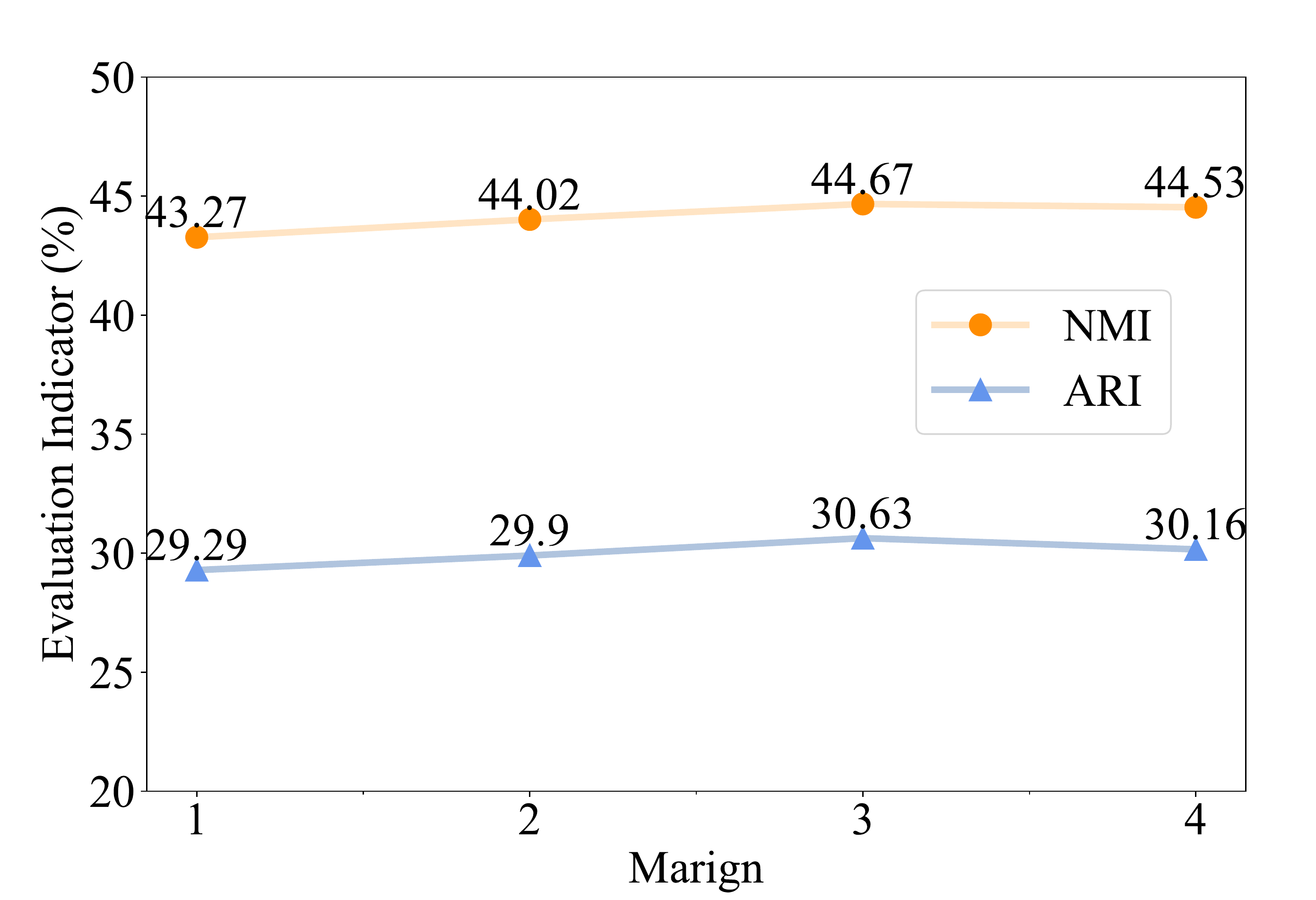}
		\label{fig53}
	\end{minipage}
	\caption{(a) Shen F value on the Movie Dialogue dataset with different session numbers (left). (b) The influence of margin m in equation \ref{eq:marign} (right). Zoom in for the best view.}
	\label{clusternum}
\end{figure}

\textbf{Session Number}. At this stage, we conduct experiments on the Movie Dialogue dataset to investigate how the session number in a dataset will influence the performance of models. We compare the Shen F value of E2E, CluCDD w/o SFF and CluCDD trained on different settings of session number. Since the Movie Dialogue dataset only occupies dialogues with 2, 3 or 4 sessions, we split the Movie Dialogue dataset into three subsets according to the number of sessions in one dialogue. We train and evaluate models on different subsets, and the result is shown in Fig \ref{clusternum}a. We can see that our CluCDD outperforms baselines in different numbers of sessions.

\noindent \textbf{Influence of margin}. We further conduct experiments on the Movie Dialogue dataset to study how our model is influenced by the contrastive loss margin $m$. In equation \ref{eq:marign}, the margin defines a radius, and antagonistic pairs contribute to the loss function only if their distance is within this radius. Meanwhile, the margin can keep contrastive loss from dropping below zero. As Fig \ref{clusternum}b reveals, the value of margin affects the performance of our model limitedly, which proves that our model maintains the performance in a wide range of margins. 

\noindent \textbf{Comparision of clustering methods}.
\begin{table}
	\centering
	\setlength{\tabcolsep}{1mm}
	\caption{Comparison about clustering methods}
	\begin{tabular}{lllll}
		\toprule
		Experiments  & NMI &  ARI  & Lo$\textbf{c}_{3}$ & Shen-F\\
		\midrule
		CluCDD+K-means & 40.98  &  31.45 & 67.98 & 67.92 \\
		CluCDD+GMM & 39.52 & 29.72 & 65.99 & 67.94 \\
		CluCDD+DBSCAN  & 39.97 & 23.12 & 65.27 & 62.76 \\
		CluCDD+AP   & 46.02  & 29.06 & 68.37 & 65.62 \\
		\bottomrule
	\end{tabular}
	
	\label{tab:clustering}
\end{table}
To further investigate the performance of different clustering methods, we compare K-means~\cite{kmeans}, Gaussian mixtures model (GMM)~\cite{GMM}, DBSCAN~\cite{DBSCAN} and Affinity propagation (AP)~\cite{AP}. 

K-means is the most common partition-based clustering method, which aims to partition n samples into k clusters in which each sample belongs to the cluster with the nearest distance. GMM is a probabilistic model for representing the presence of subpopulations within an overall population, which also separates n samples into k clusters. DBSCAN is a dense-based clustering method that directly searches for connected dense regions in the feature space by estimating the density. AP does not require the number of clusters to be determined or estimated, which also finds representative exemplars of several clusters. 

The result is shown in Table \ref{tab:clustering}. The Gaussian mixtures model performs a little worse than K-means. Meanwhile, DBSCAN is not as good as the three other methods, which is due to DBSCAN relying on two parameters: neighborhood size in terms of distance and the minimum number of points in a neighborhood. Mainly, there exist gaps between different dialogues, for which it's tough to search parameters for DBSCAN in all dialogues. The satisfactory result of Affinity propagation proves that the similarity between two utterances generated by our CluCDD is suitable for exploring the best cluster center, even without relying on the cluster number. 

\section{Conclusion}

In this work, we introduce an effective method for dialogue disentanglement. Our method is motivated by a general assumption that clustering utterances through contrastive learning. Based on this assumption, we propose a contrastive framework by distinguishing the difference between utterances in different sessions. We employ contrastive training and cluster head to construct the utterances feature space to fit the final clustering process. On average, the encouraging experimental results demonstrate that our method outperforms previous methods by 15.7\% on the Movie Dialogue dataset and 28.2\% on the Ubuntu IRC dataset.  

\newpage
\bibliographystyle{IEEEbib}
\bibliography{strings,refs}

\begin{thebibliography}{10}

\bibitem{lowe-etal-2015-ubuntu}
Ryan Lowe, Nissan Pow, Iulian Serban, and Joelle Pineau,
\newblock ``The {U}buntu dialogue corpus: A large dataset for research in
  unstructured multi-turn dialogue systems,''
\newblock in {\em Proceedings of the 16th Annual Meeting of the Special
  Interest Group on Discourse and Dialogue}, Prague, Czech Republic, Sept.
  2015, pp. 285--294, Association for Computational Linguistics.

\bibitem{dad.2017.102}
Ryan Lowe, Nissan Pow, Iulian~Vlad Serban, Laurent Charlin, Chia-Wei Liu, and
  Joelle Pineau,
\newblock ``{Training End-to-End Dialogue Systems with the Ubuntu Dialogue
  Corpus},''
\newblock {\em Dialogue \& Discourse}, vol. 8, no. 1, pp. 31--65, 2017.

\bibitem{jia-etal-2020-multi}
Qi~Jia, Yizhu Liu, Siyu Ren, Kenny Zhu, and Haifeng Tang,
\newblock ``Multi-turn response selection using dialogue dependency
  relations,''
\newblock in {\em Proceedings of the 2020 Conference on Empirical Methods in
  Natural Language Processing (EMNLP)}, Online, Nov. 2020, pp. 1911--1920,
  Association for Computational Linguistics.

\bibitem{mpcsuvey}
Jia{-}Chen Gu, Chongyang Tao, and Zhen{-}Hua Ling,
\newblock ``Who says what to whom: {A} survey of multi-party conversations,''
\newblock in {\em Proceedings of the Thirty-First International Joint
  Conference on Artificial Intelligence, {IJCAI} 2022, Vienna, Austria, 23-29
  July 2022}, Luc~De Raedt, Ed. 2022, pp. 5486--5493, ijcai.org.

\bibitem{ouchi2016addressee}
Hiroki Ouchi and Yuta Tsuboi,
\newblock ``Addressee and response selection for multi-party conversation,''
\newblock in {\em Proceedings of the 2016 Conference on Empirical Methods in
  Natural Language Processing}, 2016, pp. 2133--2143.

\bibitem{mehri-carenini-2017-chat}
Shikib Mehri and Giuseppe Carenini,
\newblock ``Chat disentanglement: Identifying semantic reply relationships with
  random forests and recurrent neural networks,''
\newblock in {\em Proceedings of the Eighth International Joint Conference on
  Natural Language Processing (Volume 1: Long Papers)}, Taipei, Taiwan, Nov.
  2017, pp. 615--623, Asian Federation of Natural Language Processing.

\bibitem{CISIR}
Jyun-Yu Jiang, Francine Chen, Yan-Ying Chen, and Wei Wang,
\newblock ``Learning to disentangle interleaved conversational threads with a
  siamese hierarchical network and similarity ranking,''
\newblock in {\em Proceedings of the 2018 Conference of the North American
  Chapter of the Association for Computational Linguistics: Human Language
  Technologies, Volume 1 (Long Papers)}, 2018, pp. 1812--1822.

\bibitem{ijcai2020-535}
Hui Liu, Zhan Shi, Jia-Chen Gu, Quan Liu, Si~Wei, and Xiaodan Zhu,
\newblock ``End-to-end transition-based online dialogue disentanglement,''
\newblock in {\em Proceedings of the Twenty-Ninth International Joint
  Conference on Artificial Intelligence, {IJCAI-20}}, Christian Bessiere, Ed. 7
  2020, pp. 3868--3874, International Joint Conferences on Artificial
  Intelligence Organization,
\newblock Main track.

\bibitem{liu2021unsupervised}
Hui Liu, Zhan Shi, and Xiaodan Zhu,
\newblock ``Unsupervised conversation disentanglement through co-training,''
\newblock in {\em Proceedings of the 2021 Conference on Empirical Methods in
  Natural Language Processing}, 2021, pp. 2345--2356.

\bibitem{devlin-etal-2019-bert}
Jacob Devlin, Ming-Wei Chang, Kenton Lee, and Kristina Toutanova,
\newblock ``{BERT}: Pre-training of deep bidirectional transformers for
  language understanding,''
\newblock in {\em Proceedings of the 2019 Conference of the North {A}merican
  Chapter of the Association for Computational Linguistics: Human Language
  Technologies, Volume 1 (Long and Short Papers)}, Minneapolis, Minnesota, June
  2019, pp. 4171--4186, Association for Computational Linguistics.

\bibitem{kummerfeld-etal-2019-large}
Jonathan~K. Kummerfeld, Sai~R. Gouravajhala, Joseph~J. Peper, Vignesh Athreya,
  Chulaka Gunasekara, Jatin Ganhotra, Siva~Sankalp Patel, Lazaros~C
  Polymenakos, and Walter Lasecki,
\newblock ``A large-scale corpus for conversation disentanglement,''
\newblock in {\em Proceedings of the 57th Annual Meeting of the Association for
  Computational Linguistics}, Florence, Italy, July 2019, pp. 3846--3856,
  Association for Computational Linguistics.

\bibitem{zhang2021discovering}
Hanlei Zhang, Hua Xu, Ting{-}En Lin, and Rui Lyu,
\newblock ``Discovering new intents with deep aligned clustering,''
\newblock 2021, pp. 14365--14373, {AAAI} Press.

\bibitem{LSTM}
Alex Graves,
\newblock ``Long short-term memory,''
\newblock {\em Supervised sequence labelling with recurrent neural networks},
  pp. 37--45, 2012.

\bibitem{kmeans}
J.~McQueen,
\newblock ``Some methods for classification and analysis of multivariate
  observations,''
\newblock {\em Computer and Chemistry}, vol. 4, pp. 257--272, 01 1967.

\bibitem{hadsell2006dimensionality}
Raia Hadsell, Sumit Chopra, and Yann LeCun,
\newblock ``Dimensionality reduction by learning an invariant mapping,''
\newblock in {\em 2006 IEEE Computer Society Conference on Computer Vision and
  Pattern Recognition (CVPR'06)}. IEEE, 2006, vol.~2, pp. 1735--1742.

\bibitem{GMM}
Carl~Edward Rasmussen et~al.,
\newblock ``The infinite gaussian mixture model.,''
\newblock in {\em NIPS}. Citeseer, 1999, vol.~12, pp. 554--560.

\bibitem{DBSCAN}
Martin Ester, Hans-Peter Kriegel, J{\"o}rg Sander, Xiaowei Xu, et~al.,
\newblock ``A density-based algorithm for discovering clusters in large spatial
  databases with noise.,''
\newblock in {\em kdd}, 1996, vol.~96, pp. 226--231.

\bibitem{AP}
Kaijun Wang, Junying Zhang, Dan Li, Xinna Zhang, and Tao Guo,
\newblock ``Adaptive affinity propagation clustering,''
\newblock {\em arXiv preprint arXiv:0805.1096}, 2008.

\end{thebibliography}

\end{document}